\documentclass[letterpaper, 10 pt, journal, twoside]{IEEEtran}

\usepackage{amsmath,amsfonts}
\usepackage{algorithmic}
\usepackage{algorithm}
\usepackage{array}
\usepackage[caption=false,font=normalsize,labelfont=sf,textfont=sf]{subfig}
\usepackage{textcomp}
\usepackage{stfloats}
\usepackage{url}
\usepackage{verbatim}
\usepackage{graphicx}
\usepackage{cite}
\usepackage{adjustbox}
\usepackage{xcolor}
\usepackage{soul}
\definecolor{lightpink}{rgb}{1.0, 0.71, 0.76}
\definecolor{lightcoral}{rgb}{1.0, 0.75, 0.5}
\definecolor{lavender}{rgb}{0.71, 0.53, 0.83}
\usepackage{gensymb}
\usepackage{makecell}
\usepackage{bbding}
\newcolumntype{P}[1]{>{\raggedright\arraybackslash}p{#1}}

\begin{document}

\title{
Enabling High-Curvature Navigation in Eversion Robots through Buckle-Inducing Constrictive Bands 


}





\author{Cem Suulker$^{1}$, Muhie Al Haimus$^{1}$, Thomas Mack$^{1}$, Mohammad Sheikhsofla$^{1}$, Neri Niccolò Dei$^{1}$, Reza Kashef$^{1}$, Hadi Sadati$^{1}$, Federica Barontini$^{2}$, Fanny Ficuciello$^{3}$, Alberto Arezzo$^{2}$, Bruno Siciliano$^{3}$~\IEEEmembership{Fellow,~IEEE}, Sebastien Ourselin$^{4}$, and Kaspar Althoefer$^{1}$~\IEEEmembership{Fellow,~IEEE}

\thanks{This work has been submitted to the IEEE for possible publication. Copyright may be transferred without notice, after which this version may no longer be accessible.}
\thanks{This work is supported by ERC grant EndoTheranostics, 101118626. Funded by the European Union.}
\thanks{Cem Suulker was funded by Ministry of National Education of Türkiye.}
\thanks{Thomas Mack is supported by an iCASE EPSRC PhD studentship with Nuclear Restoration Services (NRS-Dounreay).}
\thanks{For the purpose of open access, the author(s) has applied a Creative Commons Attribution (CC BY) license to any Accepted Manuscript version arising.}
\thanks{$^{1} $Authors are with the Centre for Advanced Robotics @ Queen Mary, School of Engineering and Materials Science, Queen Mary University of London, United Kingdom.
        {\tt\footnotesize c.suulker@qmul.ac.uk}}
\thanks {$^{2} $ Authors are with Department of Surgical Sciences, University of Turin, Città della Salute e della Scienza di Torino, Turin, Italy.}
\thanks {$^{3} $ Author is with the PRISMA Lab, Department of Electrical Engineering and Information Technology, University of Naples Federico II, 80125 Naples, Italy.}
\thanks {$^{4} $ Author is with School of Biomedical Engineering and Imaging Sciences, King’s College London, London, United Kingdom.}
}



\maketitle

\begin{abstract}

Tip-growing eversion robots are renowned for their ability to access remote spaces through narrow passages. However, achieving reliable navigation remains a significant challenge. Existing solutions often rely on artificial muscles integrated into the robot body or active tip-steering mechanisms. While effective, these additions introduce structural complexity and compromise the defining advantages of eversion robots: their inherent softness and compliance. In this paper, we propose a passive approach to reduce bending stiffness by purposefully introducing buckling points along the robot's outer wall. We achieve this by integrating inextensible diameter-reducing circumferential bands at regular intervals along the robot body facilitating forward motion through tortuous, obstacle cluttered paths. Rather than relying on active steering, our approach leverages the robot's natural interaction with the environment, allowing for smooth, compliant navigation. We present a Cosserat rod–based mathematical model to quantify this behavior, capturing the local stiffness reductions caused by the constricting bands and their impact on global bending mechanics. Experimental results demonstrate that these bands reduce the robot's stiffness when bent at the tip by up to 91\%, enabling consistent traversal of 180\degree\ bends with a bending radius of as low as 25 mm---notably lower than the 35 mm achievable by standard eversion robots under identical conditions. The feasibility of the proposed method is further demonstrated through a case study in a colon phantom. By significantly improving maneuverability without sacrificing softness or increasing mechanical complexity, this approach expands the applicability of eversion robots in highly curved pathways, whether in relation to pipe inspection or medical procedures such as colonoscopy.
\end{abstract}


\begin{figure}[t]
  \centering
  \includegraphics[width=1\linewidth]{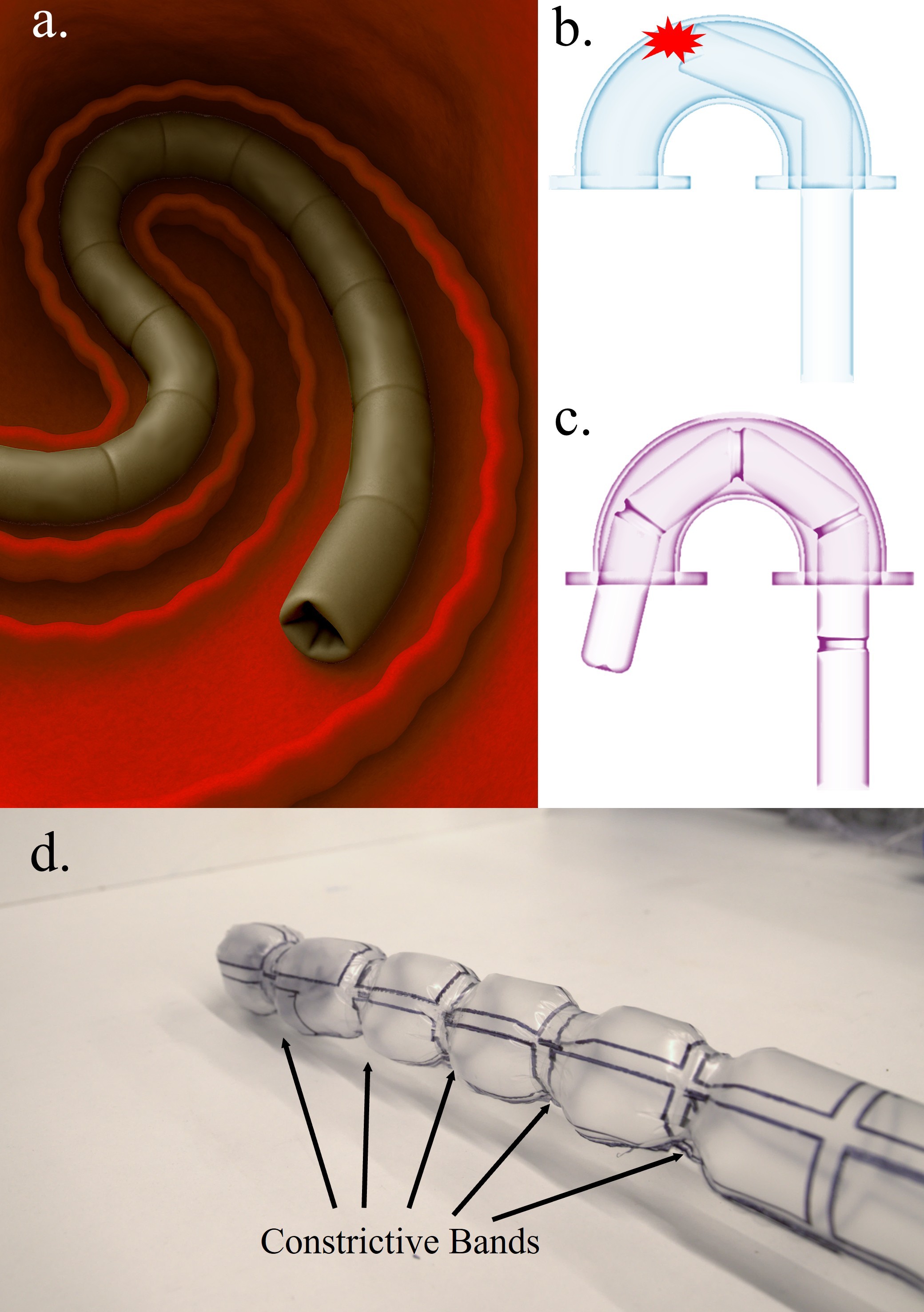}
  \caption{a) Concept drawing of an eversion robot equipped with constrictive bands navigating through a colon-like environment. b) Illustration of a standard eversion robot getting stuck at a sharp bend. c) Illustration of the proposed eversion robot with integrated constrictive bands successfully navigating the same bend. d) Detailed view of an eversion robot with the constrictive bands incorporated.}
  \label{fig1}
\end{figure}

\section{Introduction}

\begin{table*}[h]
    \centering
    \renewcommand{\arraystretch}{1.4}
    \setlength{\tabcolsep}{6pt}
        \caption{Comparison of navigation methods with respect to their features. \Checkmark indicates feature is satisfied, X not satisfied.}
    \begin{tabular}{c c c c c c c c }
        \hline
        \textbf{Paper} & \textbf{Year} & \makecell{\textbf{Navigation} \\ \textbf{Method}} & \textbf{Passive} & \makecell{\textbf{Path knowledge} \\ \textbf{not needed}} & \makecell{\textbf{Soft \&} \\ \textbf{squeezable}} & \makecell{\textbf{No additional} \\ \textbf{friction}} & \makecell{\textbf{No extra} \\ \textbf{material layers}} \\ \hline
         \cite{coad2019vine} & 2019 & Attached PAMs & X & \Checkmark & \Checkmark & \Checkmark & X \\ \hline
        \cite{abrar2021highly} & 2021 & Integrated PAMs & X & \Checkmark & \Checkmark & \Checkmark & X \\ \hline
         \cite{haggerty2021hybrid} & 2021 & Tip steering with internal mechanism & X & \Checkmark & X & \Checkmark & \Checkmark \\ \hline
        \cite{al2024hybrid} & 2024 & Tip steering with help of cap & X & \Checkmark & X & X & \Checkmark \\ \hline
        \cite{greer2020robust} & 2020 & Pre-shaped bending & \Checkmark & X & \Checkmark & \Checkmark & \Checkmark \\ \hline
         \textbf{This work} & 2025 & Constrictive bands & \Checkmark & \Checkmark & \Checkmark & \Checkmark & \Checkmark \\ \hline
    \end{tabular}
    \label{tab:navigation_features}
\end{table*}

\IEEEPARstart{E}{version} robots are a class of soft growing robots that extend from their tip in a manner reminiscent of growing vines. Also referred to as vine robots, they are operated via fluidic actuation and are typically constructed from flexible sheet materials such as thin plastic films or coated fabrics \cite{ourreview}. Initially, the material is folded into itself, and when pressurized, it unfolds from the inside out, propelling the robot forward from the tip. These robots can grow to many times their original length, and are ideally suited to tasks requiring access to confined or hard-to-reach environments. Their compliant structure ensures minimal interaction force with the surrounding environment, a characteristic that is particularly advantageous in sensitive medical procedures, such as colonoscopy.

In constrained environments like tubing systems or anatomical lumens, eversion robots tend to follow the path of least resistance---unless faced with obstacles or sharp turns. Conventional eversion robots have been shown to accommodate bends up to a certain threshold (180\degree, more than 45.1 mm bending radius, at 3.5 kPa in the work by Haggerty et al. \cite{haggerty2019characterizing}) before getting stuck. In medical applications such as colonoscopy, however, the robot needs to reliably navigate multiple bends with much smaller bending radii \cite{ourreview}---a feat beyond the capacity of standard eversion designs.

To address these limitations, various navigation strategies have been explored. Broadly, these can be classified into active and passive steering mechanisms. Active methods incorporate additional actuators to dynamically adjust the robot shape and trajectory during operation. These approaches, such as tip steering using rigid elements \cite{haggerty2021hybrid}, or the integration of pneumatic artificial muscles (PAMs) along the body \cite{coad2019vine, abrar2021highly, mack2025efficient} , all offer precise control but introduce added design complexity. Typically, this increases overall stiffness, compromising key benefits such as smooth movement(due to additional unfolding layers) and squeezability (due to the introduction of rigid elements).

While passive approaches are less intrusive, they typically involve setting fixed predetermined configurations that dictate the robot’s curvature and are unsuitable for environments with unknown geometries \cite{greer2020robust, agharese2023configuration}. Thus, there remains a need for simple, passive navigation enhancement strategies that maintain the robot’s soft and compliant nature while improving its ability to conform to sharp turns (Table \ref{tab:navigation_features}).

In this paper, we propose a structural enhancement for eversion robots by integrating diameter-reducing constrictive bands into the robot body. This reduces  bending stiffness by inducing controlled buckling zones, ultimately improving passive navigation performance (Figure \ref{fig1}d). The deformation of the eversion robot is modeled and analyzed mathematically by treating it as a Cosserat rod whose cross-section exhibits the mechanics of a pressurized thin-walled tube. The analysis reveals that the bands not only modify the geometric profile but also significantly alter the mechanical properties of the tube. These bands are strategically integrated during fabrication and allow the robot to bend more easily without compromising its structural integrity or unfolding behavior. This method retains the essential advantages of eversion-based motion while enabling reliable navigation in constrained environments with sharp bends.

The remainder of this paper is structured as follows: Section II mathematically models the buckling motion caused by the introduction of the bands. Section III details the materials and fabrication process, while Section IV presents experimental methodology and results evaluating stiffness reduction, enhanced bend performance, and any trade-offs such as impact on eversion smoothness. Section V details a case study involving a colon phantom that demonstrates practical benefits in a representative medical setting. Section VI offers a discussion of the findings and concludes the paper.



\section{Mathematical Model of Deformation}
To analyze the deformation behavior, the eversion robot is modeled as a Cosserat rod whose cross-section corresponds to a pressurized thin-walled tube. The configuration of the tube is described by the Cartesian position of its centerline, $\boldsymbol{P}(s) \in \mathbb{R}^3$, and the rotation matrix of its material frame, $\boldsymbol{R}(s) \in SO(3)$, both expressed as functions of the arc length $s \in [0,L]$, where $L$ denotes the total length of the tube. To describe the static equilibrium of the elastic tube subjected to an external contact force $\boldsymbol{F}$  applied at the arc-length $s=s_c$, the Cosserat rod model \cite{till2019real, sofla2024haptic} is formulated as the following system of ordinary differential equations:
\begin{subequations}
\begin{align}
\dot{\boldsymbol{R}} &= \boldsymbol{R} \left( \mathbf{K}_{bt}^{-1} \boldsymbol{R}^T \boldsymbol{m} \right)^\wedge, \\
\dot{\boldsymbol{P}} &= \boldsymbol{R} \left( \mathbf{K}_{se}^{-1} \boldsymbol{R}^T \boldsymbol{n} + \begin{bmatrix} 0 & 0 & 1 \end{bmatrix}^T \right), \\
\dot{\boldsymbol{n}} &= -\boldsymbol{F} \, \delta(s-s_c),\\
\dot{\boldsymbol{m}} &= -\dot{\boldsymbol{P}} \times \boldsymbol{n},
\end{align}
\end{subequations}

where the dot denotes differentiation with respect to $s$, the operator $(\cdot)^\wedge$ maps a vector in $\mathbb{R}^3$ to its skew-symmetric matrix in $\mathfrak{so}(3)$, and $\delta(\cdot)$ is the Dirac delta function. The vectors $\boldsymbol{n}(s)$ and $\boldsymbol{m}(s)$ represent the internal force and moment along the rod, respectively. The constitutive stiffness matrices for shear/extension ($\mathbf{K}_{se}$) and bending/torsion ($\mathbf{K}_{bt}$) are defined as:

\begin{equation}
\mathbf{K}_{se} = A
\begin{bmatrix}
G & 0 & 0 \\
0 & G & 0 \\
0 & 0 & E
\end{bmatrix}, 
\quad
\mathbf{K}_{bt} =
\begin{bmatrix}
EI & 0 & 0 \\
0 & EI & 0 \\
0 & 0 & 2GI
\end{bmatrix}.
\label{eq:stiffness}
\end{equation}

where $E$ and $G$ are the Young's modulus and the shear modulus, respectively, $A$ is the cross-sectional area, and $I$ is the geometric moment of inertia of the cross-section. Because the tube diameter decreases at the sections with bands, both $A(s)$ and $I(s)$ are treated as functions of the arc length $s$.

To incorporate the influence of internal pressure on structural rigidity, the tube is modeled as a pressurized thin shell exhibiting pressure-dependent bending resistance. Internal pressure induces circumferential and axial membrane stresses, which increase the apparent bending stiffness of the structure. This pressure-stiffening behavior is represented by an effective elastic modulus $E_{\mathrm{eff}}(p)$, which encapsulates both material elasticity and geometric stiffening due to membrane tension.

When diameter-reduced (constricted) segments are included in the simulation, measured deflections exceed those predicted by the homogeneous Cosserat model. This discrepancy arises because the sections with bands violate several modeling assumptions: they introduce geometric discontinuities and local loss of membrane pre-tension (wrinkling). These effects reduce the pressure-induced geometric stiffening that the continuum model assumes. To account for this, a local reduction factor $\alpha(s) \in (0,1]$ is introduced, and the effective modulus is defined as:

\begin{equation}
E(p,s) = \alpha(s) \, E_{\mathrm{eff}}(p).
\label{eq:effective_modulus}
\end{equation}

This correction allows the Cosserat model to reflect the locally weakened bending resistance caused by the constrictive bands, thereby improving agreement with experimental observations.


In the following sections, we validate our mathematical findings through experimental analysis. To accomplish this, we fabricated eversion robot prototypes with inextensible constrictive bands.

\section{Materials and Fabrication}

This section provides details of the materials used in, and an explanation of the fabrication process of, our prototype robots. 

\begin{figure}[ht]
  \centering
  \includegraphics[width=1\linewidth]{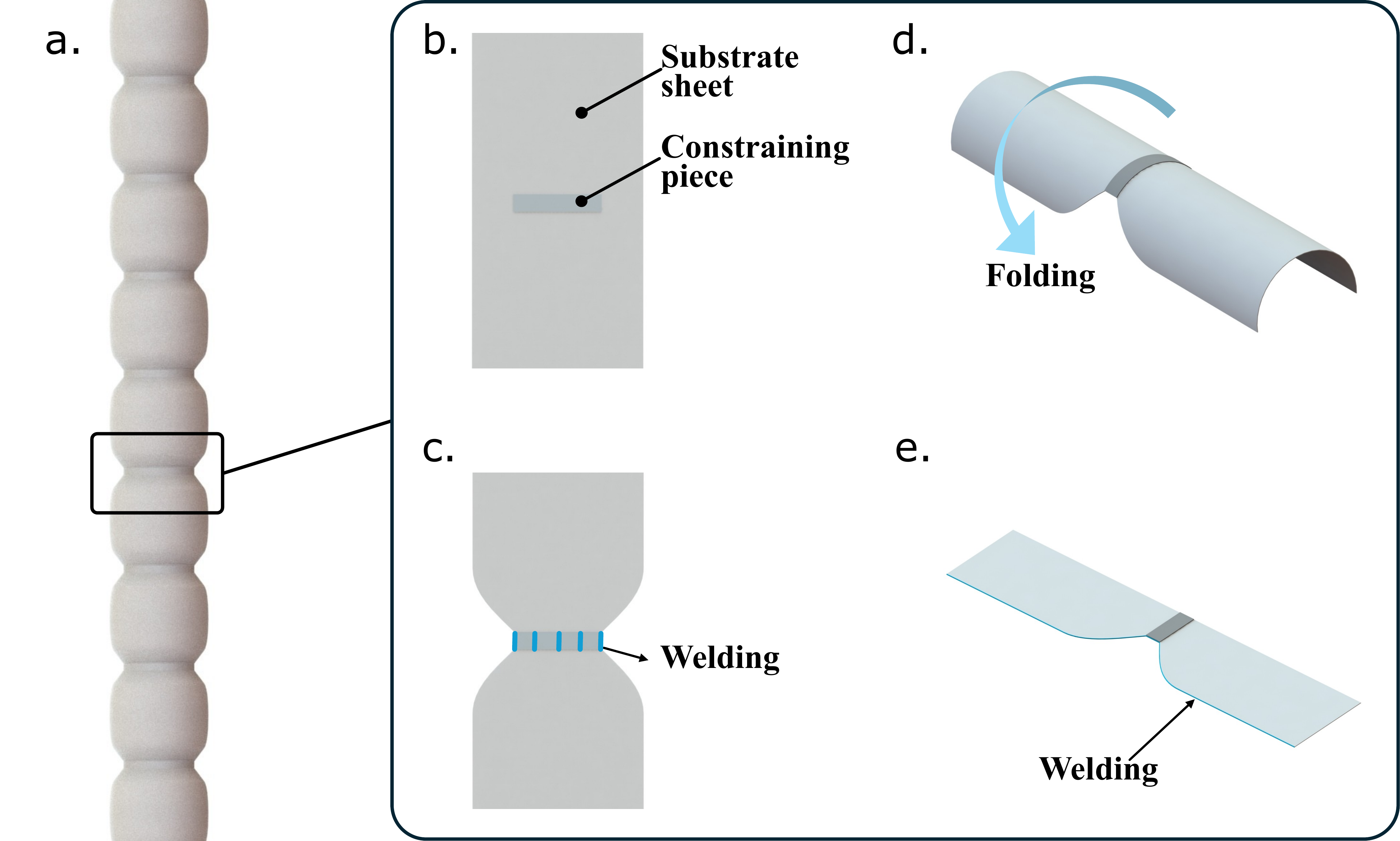}
  \caption{Outline of the fabrication process with constrictive band implementation. a) Overview of the eversion robot with constrictive bands. b) The eversion robot is manufactured using a substrate sheet while each constraining band is made of a further piece of the same TPU material. c) The ends of each constraining piece are sealed to the left and right edges of the substrate sheet, while three intermediate points are also affixed to ensure uniform contraction and circular symmetry during eversion. Sealing is achieved using an ultrasonic welding machine. d) Once the band strips are fixed into place, the sheet is carefully folded lengthwise to align its edges, creating a tubular form. e) The edges are then sealed along the seam to create an airtight cylindrical structure, later to be inverted to create an eversion robot.}
  \label{Fabrication}
\end{figure}

For this study, thermoplastic polyurethane (TPU) with a thickness of 0.05 mm (DOONX HFU120) was selected as the base material for the eversion robot body. A customizable, high precision Vetron 5064 ultrasonic welder---which functions much like a sewing machine---was used to seal the TPU sheets. All robot prototypes were manufactured with a diameter of 40 mm and a length of 600 mm for consistency. Three prototypes were made for each robot variant, differing only in the band placement positions and the degree of diameter reduction.

Fabrication began by laser cutting TPU sheets into pieces measuring 125.7 mm in width and 600 mm in length. As the same TPU material was used for the constrictive reducing bands, 15 mm wide variable length strips were laser cut per the required reductions. For example, to reduce the diameter from 40 mm to 20 mm, a TPU strip of 62.8 mm was cut (calculated using $\pi \times 20 = 62.8$).

These TPU strips were positioned at the designated diameter-reduction points (Figure \ref{Fabrication}b). The ends of each strip were ultrasonic welded to the corresponding ends of the TPU sheet, with three additional attachment points spaced evenly along the strip to ensure uniform diameter reduction (Figure \ref{Fabrication}c). The sheet was then folded onto itself, and the ultrasonic welder guided along the edge to form a cylindrical tube (Figures \ref{Fabrication}d, and e), which was later inverted to create the eversion robot.


Manufacturing steps were carried out using a 3D-printed alignment mask, which enabled high-precision marking for subsequent ultrasonic welding. This approach ensured that the prototypes were fabricated with high repeatability and accuracy.

If a fabric-based material is selected instead of TPU, the same fabrication steps apply, with modifications to suit the textile assembly. In such cases, a sewing machine is used in place of an ultrasonic welder. The band can be attached using the ruffling method, optionally with an elastic band, as demonstrated by \cite{suulker2022soft, suulker2024let}. Alternatively, hand-stitching using a needle and thread is a method that offers finer control. An overlocker can be used to finish the seams and complete the cylindrical robot structure. Lastly, the robot needs to be sealed using brushed latex, vinyl tape, or by bonding with TPU.



\section{Experimental Characterization and Validation}

The contribution of this work is the introduction of constrictive bands that intentionally induce programmed buckling along the eversion robot body. By locally constraining the diameter, these bands significantly reduce overall bending stiffness, thereby enhancing the robot’s ability to negotiate sharper bends without adding mechanical complexity. To rigorously evaluate this effect, we designed a comprehensive set of characterization experiments to quantify stiffness reduction and elucidate the underlying mechanisms. The experimental results were plotted with mathematical simulations performed in MATLAB, and showed clear agreement and mutual validation. Finally, a validation experiment was conducted to assess the impact of the proposed design on navigation performance in confined environments, confirmed that the constrictive bands do  improve passive maneuverability.


All tests were conducted using three independently fabricated robot prototypes per configuration to account for manufacturing variability and to increase the reliability of the results. For experiments that did not require eversion, the constriction was achieved using cable ties with accurately adjustable diameters. This ensured precise diameter restriction and eliminated inconsistencies introduced by the band integration. All reported measurements are presented as mean values with standard deviations. The specifications of the tested prototypes are summarized in Table \ref{tab:samples}.

\begin{table}[h]
    \centering
        \caption{Characteristics of each prototype and their corresponding bending stiffness measurements from experimental studies. Each prototype was fabricated in triplicate, and all experiments were repeated three times per prototype (four times for the validation experiment). Stiffness values were provided via the means and standard deviations for each of these readings.}
    \begin{tabular}{{P{0.9cm} P{1cm} P{1.25cm} P{1.85cm} P{1.55cm}}}
        \hline
        Prot. \# & Band \# & Diameter reduction & Band placement from tip (mm) & B. Stiffness I. $N/m$ (SD) \\ \hline
        \multicolumn{4}{l}{\textbf{Characterization Experiment 1}} \\ \hline
        1 & 0 & N/A & N/A & 11.94 (0.42) \\
        2 & 1 & 50\% & 50 & 7.75 (0.28) \\
        3 & 2 & 50\% & 50,100 & 3.20 (0.19)  \\
        4 & 3 & 50\% & 50,100,150 & 1.92 (0.04)  \\
        5 & 4 & 50\% & 50,100,150,200 & 1.11 (0.04)  \\ \hline
        \multicolumn{4}{l}{\textbf{Characterization Experiment 2}} \\ \hline
        1 & 0 & N/A & N/A & 11.94 (0.42)  \\
        2 & 1 & 50\% & 30 & 9.52 (0.09)  \\
        3 & 1 & 50\% & 40 & 8.77 (0.06)  \\
        4 & 1 & 50\% & 50 & 7.75 (0.28)  \\ 
        5 & 1 & 50\% & 60 & 7.33 (0.39)  \\ 
        6 & 1 & 50\% & 70 & 6.66 (0.95) \\
        7 & 1 & 50\% & 80 & 5.18 (0.29)  \\
        8 & 1 & 50\% & 90  & 4.47 (0.14) \\
        9 & 1 & 50\% & 100 & 3.81 (0.56)  \\  \hline
        \multicolumn{4}{l}{\textbf{Characterization Experiments 3 and 4}} \\ \hline
        1 & 0 & N/A & N/A & 11.94 (0.42)  \\
        2 & 1 & 10\% & 100 & 9.92 (0.04) \\
        3 & 1 & 20\% & 100 & 9.51 ($<$0.01)  \\
        4 & 1 & 30\% & 100 & 8.80 (0.43)  \\
        5 & 1 & 40\% & 100 & 6.96 (0.13)  \\ 
        6 & 1 & 50\% & 100 & 3.81 (0.4)  \\ \hline
        \multicolumn{4}{l}{\textbf{Validation Experiment}} \\ \hline
        1 & 0 & N/A & N/A & N/A \\
        2 & 3 & 10\% & 50 mm gaps & N/A \\
        3 & 3 & 20\% & 50 mm gaps & N/A \\
        4 & 3 & 30\% & 50 mm gaps & N/A  \\ \hline
    \end{tabular}
    \label{tab:samples}
\end{table}

\subsection{Characterization Experiment 1: Effect of Number of Bands on Stiffness}

The goal of this experiment is to determine whether the introduction of constrictive bands leads to a reduction in the bending stiffness of the tip of the eversion robot and, furthermore, whether increasing the number of bands enhances this effect. To evaluate this, we designed the following experimental protocol.

\subsubsection{Experimental Protocol}
An Instron 5967 testing machine equipped with a 2525-800 Series 100 N load cell ($\pm0.25$\%) was used for this experiment. The setup allowed the machine to pull on the tip of the robot while simultaneously recording both the resisting force and the displacement of the robot tip.

\begin{figure*}[htp]
  \centering
  \includegraphics[width=1\linewidth]{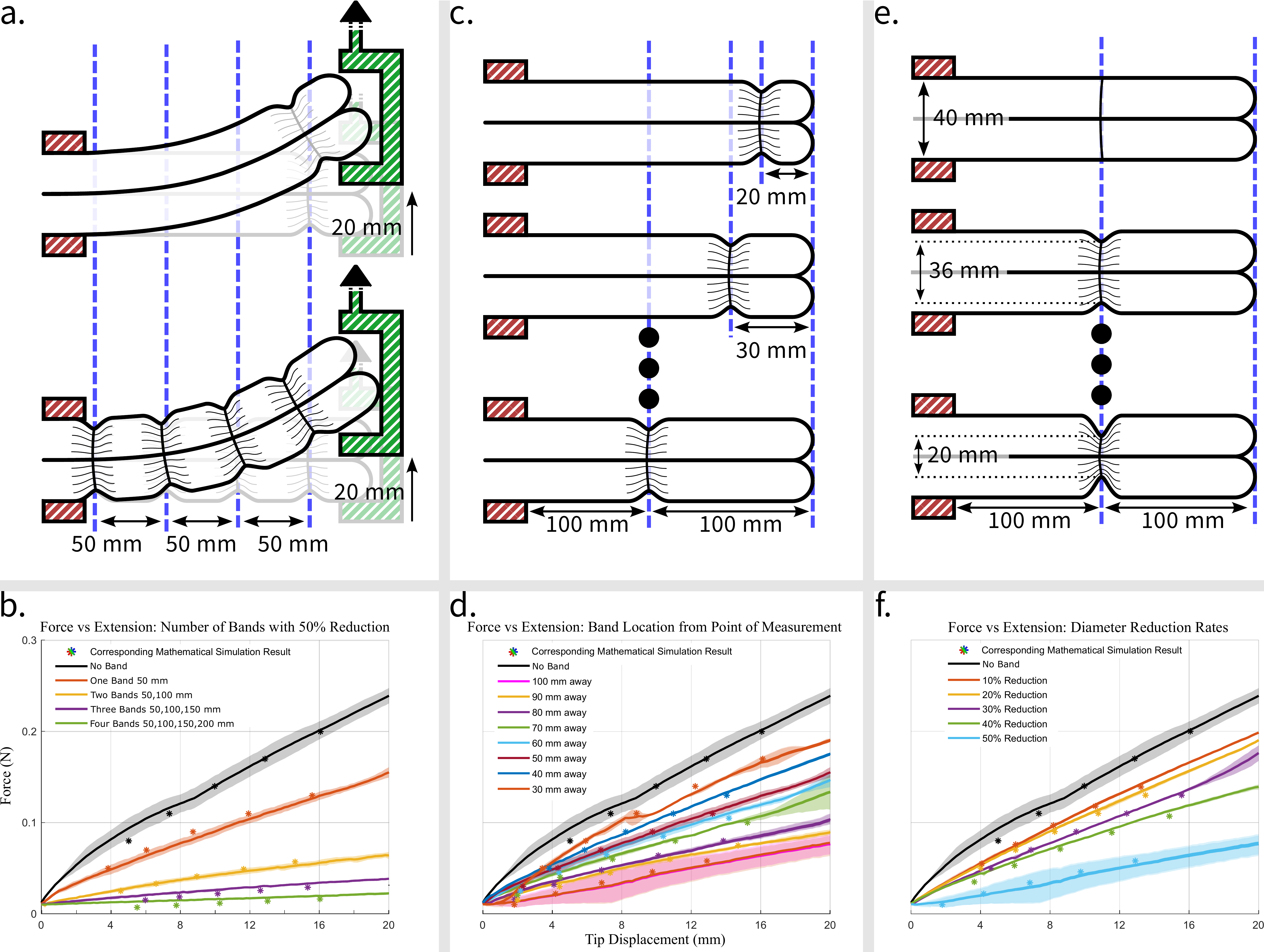}
\caption{Experimental configurations and results for characterization experiments 1–3.  In all the experiments, the eversion robot was displaced laterally from the tip by 20 mm. The mathematical simulation results are scattered on plots. a) In Experiment 1, the number of constrictive bands was varied, and the corresponding results are shown in b). c) In Experiment 2, the band position was altered, and these results are shown in d). e) In Experiment 3, the diameter reduction rate was varied, and these results are presented in f).}

  \label{CharacterisationExperiments}
\end{figure*}

The robot was fully everted and inflated to a constant internal pressure of 6.9 kPa (1 psi), regulated using an SMC ITV2050 pressure regulator. Its tip was secured to the Instron testing machine and positioned such that it made only minimal contact with the force sensor, while any residual sensor noise was zeroed. During the experiment, the machine steadily displaced the robot tip upward while continuously recording the applied force and corresponding displacement. Each trial was terminated once the robot had deformed by 20 mm. A schematic of the experimental setup is shown in Figure \ref{CharacterisationExperiments}a.

Five robot configurations were tested: a standard robot with no bands, a robot with one band, and three others, with two, three and four bands respectively.

Each band imposed a 50\% diameter reduction. To minimize the influence of band placement, bands were evenly distributed along the robot body. In the one-band configuration, the band was  positioned 50 mm from the tip. In the two-band configuration, bands were positioned 50 mm and 100 mm from the tip; for three bands, the intervals were 50 mm, 100 mm and 150 mm; and for four bands, 50 mm, 100 mm, 150 mm and 200 mm. The experimental set-ups, the simulation results and the measured data are presented in Table \ref{tab:samples}.

\subsubsection{Results}





Force–displacement data are shown in Figure \ref{CharacterisationExperiments}b. 
The solid lines represent the mean values of the experimental measurements, while the shaded regions denote the corresponding standard deviations. The simulation results are plotted as scattered points and show consistency with the experimentally measured data. The identified effective elastic modulus for the applied internal pressure is $E_{\mathrm{eff}} = 25.2~\mathrm{MPa}$, and the local reduction factor for the 50\% diameter reduction is $\alpha = 0.36$. From the results, it can be observed that the base robot requires the highest force to deform, and that robots with integrated bands require progressively less force. As the number of bands increases, the overall stiffness of the robot decreases accordingly.  

\begin{equation} 
    \emph{k}=\dfrac{\Delta F}{\Delta x} 
    \label{eq:stiffness2} 
\end{equation}

To enable a quantitative comparison of deformation behavior, we adopted a simple stiffness index as a comparative measure of bending stiffness (Equation \ref{eq:stiffness2}). This approach is used in the characterization of soft robotic bending actuators to provide a bending stiffness metric \cite{suulker2022soft,huang2019highly}. The stiffness index of each prototype was therefore calculated using Equation \ref{eq:stiffness2}, where $\emph{k}$ is the stiffness index of the robot, 
$\Delta F$ is the measured change in force, and $\Delta x$ is the imposed displacement, set to 0.02 m (20 mm) in all relevant experiments. 
The results, summarized in Table \ref{tab:samples}, are reported as mean values with standard deviations. 
The measured stiffness index values were 11.94 N/m for the base robot, 7.75 N/m for the one-band prototype, 3.20 N/m for the two-band, 1.92 N/m for the three-band, and 1.11 N/m for the four-band version. 
This demonstrates a clear, gradual reduction in stiffness as the number of bands increases, confirming that consecutive bands provide a cumulative softening effect.  

However, it is important to note that this effect cannot be accumulated indefinitely along the entire length of the robot. 
Once the robot navigates a bend, the influence of the bands between the base and the bend location is effectively annulled for the next bend. 
Therefore, only the bands located beyond each turn will contribute to reducing the bending stiffness and therefore to enhancing the robot's ability to navigate further bends.

\subsection{Characterization Experiment 2: Effect of Band Location on Stiffness}


In addition to the number of bands, it is also important to examine how the placement of the bands influences stiffness at the robot tip. This experiment was therefore designed to evaluate the effect of band location.

\subsubsection{Experimental Protocol}



The same Instron 5967 testing machine setup was used as in the previous experiment. In this case however, the number of bands was set at one, and the diameter reduction was maintained at 50\%. The only variable was the distance of the band from the robot tip.

Nine configurations were tested: a standard robot with no bands, and robots with a single band placed at distances starting from 30 mm, and then in increments of 10 mm up to 100 mm from the tip. As before, the tip of each inflated robot was displaced upward while force and extension values were recorded (Figure \ref{CharacterisationExperiments}e).

\subsubsection{Results}



Force–displacement data from this experiment are shown with the simulation results in Figure \ref{CharacterisationExperiments}d). The simulation and experimental results indicate that as the band to tip distance increases, the force required to deflect the robot tip decreases. The base robot exhibited a bending stiffness index of 11.94 N/m, while bands positioned at 30, 40, 50, 60, 70, 80, 90, and 100 mm from the tip yielded stiffness index values of 9.52, 8.77, 7.75, 7.33, 6.66, 5.18, 4.47, and 3.81 N/m, respectively.  

These findings suggest that stiffness reduction is maximized when the constrictive band is positioned furthest from the tip. 
However, for practical applications in which the precise bending location is unknown, or indeed for any multi-bend environment, employing multiple bands along the robot body may be advantageous---all the more so, as the influence of the bands located behind any bend would diminish. 

\begin{figure*}
    \centering
    \includegraphics[width=1\linewidth]{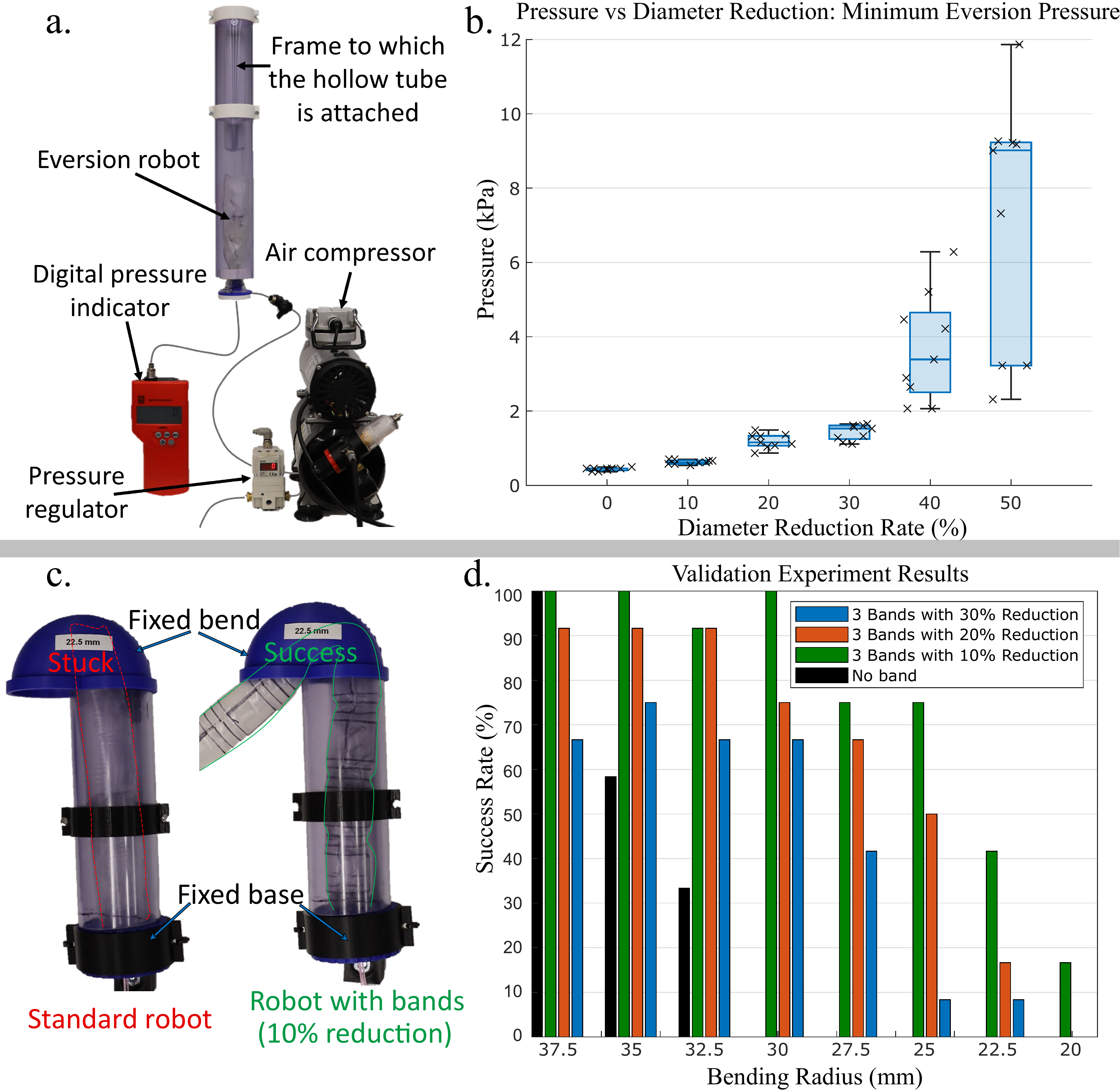}
    \caption{Top - Experimental setup and results for Characterization Experiment 4. a) The eversion robot was inflated in a controlled environment, and the minimum pressure required for full eversion was recorded. b) Results indicate a super-linear increase in the required eversion pressure as the robot diameter is increasingly restricted by the bands. Bottom - Experimental setup and results for the Validation Experiment in a confined environment. c) The robot is fixed inside a tube and inflated at 3 kPa. The distal end of the tube has 180$\degree$ turn casing attachments of progressively decreasing bending radii. The trial is deemed a success if the robot everts through the bend, and a failure if it gets stuck within it. d) The results demonstrate that the robot with 10\% constrictive bands achieves the highest overall performance in this validation experiment.}
    \label{fig:Exp4}
\end{figure*}

\subsection{Characterization Experiment 3: Effect of Diameter Reduction on Stiffness}


This experiment was conducted to investigate how varying the degree of diameter restriction influences the stiffness of the eversion robot.

\subsubsection{Experimental Protocol}



The setup was identical to that used in the previous experiments with the Instron 5967 testing machine equipped with a 2525-800 Series 100 N load cell. In this case, the diameter reduction ratio of the bands was varied.

Six different configurations were tested: a control robot without bands, and robots with a single band effecting diameter reductions of 10\%, 20\%, 30\%, 40\%, and 50\%. For consistency, all bands were positioned 100 mm from the tip of the robot. As in the previous experiment, the tip of the inflated robot was steadily displaced upward by the machine while corresponding force and extension values were recorded (Figure \ref{CharacterisationExperiments}e).

\subsubsection{Results}


Force–displacement data obtained by simulations and from experimental measurements are presented in Figure \ref{CharacterisationExperiments}f. 
The identified values of the local reduction factor ($\alpha$) for the 10\%, 20\%, 30\%, and 40\% diameter reductions are $0.55$, $0.53$, $0.46$, and $0.43$, respectively. Therefore, even a 10\% diameter reduction has a considerable effect on the effective elastic modulus, reducing it by approximately $45\%$. The results demonstrate a clear trend: as the diameter reduction introduced by the bands increases, the bending stiffness of the robot tip decreases. The base robot exhibited a bending stiffness index of 11.94 N/m. 
With diameter reductions of 10\%, 20\%, 30\%, 40\%, and 50\%, the experimental bending stiffness index values were 9.92, 9.51, 8.80, 6.96, and 3.81 N/m, respectively.  

Even a modest 10\% reduction in diameter lowered bending stiffness index by approximately 17\%. 
However, the effect was not uniform across the range: the reduction from 10\% to 20\% diameter restriction resulted in only a further 4\% decrease, 
while the difference between 40\% and 50\% reduction was much more pronounced, corresponding to a 45\% drop in the bending stiffness index. 
This indicates that the relationship is non-linear, with higher diameter reductions yielding disproportionately greater reductions in stiffness.  

These findings suggest that the most effective strategy for maximizing bending stiffness reduction is to employ the highest feasible degree of diameter reduction. However, this improvement comes at a cost: higher diameter reduction adversely affects the eversion process. Specifically, greater restriction requires increased internal pressure for eversion, and beyond a certain threshold, the eversion pressure approaches or exceeds the rupture pressure of the material. Thus, while large reductions offer significant bending stiffness benefits, they also risk limiting the robot’s operational range due to the elevated pressure demands. The following experiment examines this phenomenon in detail.





\subsection{Characterization Experiment 4: Effect of Diameter Reduction of the Bands on Minimum Eversion Pressure}

Building on the previous experiment, this test evaluated how the addition of constrictive bands affects the minimum pressure required for successful eversion.  

\subsubsection{Experimental Protocol}

The eversion robot was positioned vertically inside a controlled tubular environment, oriented to evert against gravity. Pneumatic pressure was applied gradually in increments, and the minimum pressure required to initiate eversion was recorded for each trial. Figure \ref{fig:Exp4}a illustrates the experimental setup. Pressure measurements were obtained using an RS Pro digital pressure indicator connected to the inner chamber via a channel at the robot’s base, ensuring accurate readings.  

A trial was considered successful only if the robot fully everted. If no motion was observed within 20 seconds of pressurization, the attempt was recorded as a failure and the trial was repeated with increased input pressure. Both standard robots and robots with constrictive bands of varying restriction ratios were tested and compared to assess how band geometry influences the minimum eversion pressure.

\subsubsection{Results}




The results are summarized in Figure \ref{fig:Exp4}b, shown as box plots with scattered data points for individual trials. The standard robot (0\% restriction) required a median minimum eversion pressure of 0.46 kPa (SD=0.04). For robots with diameter reductions, the required pressures were: 10\% reduction – 0.62 kPa (SD=0.06), 20\% – 1.16 kPa (SD=0.19), 30\% – 1.53 kPa (SD=0.21), 40\% – 3.39 kPa (SD=1.46), and 50\% – 9.01 kPa (SD=3.41).  

These results indicate a superlinear relationship between diameter restriction and required eversion pressure. Readings for diameter reductions of more than 50\% were not obtained as the robot with 60\% band diameter reduction failed to evert---the requisite pressure in that instance exceeding the rupture threshold of the sealing layer.  

From Figure \ref{fig:Exp4}b, it is evident that pressure increases remain relatively modest up to a 30\% restriction. Beyond this point, the required pressures rise steeply, making higher restriction ratios impractical. Therefore, for subsequent validation experiments, only the 10\%, 20\% and 30\% configurations were selected, as they provide the most favorable trade-off between bending stiffness reduction and eversion feasibility.

\subsection{Validation Experiment: Navigation Tests in Rigid Curved Paths}

In the previous experiments we have shown that the bands do indeed reduce bending stiffness of the eversion robot. This section presents experimental tests validating the claim that this novel approach facilitates navigation through sharper bends.

\subsubsection{Experimental Protocol}

A confined-path test setup was prepared (Figure \ref{fig:Exp4}c). A pipe with an internal diameter of 55 mm was used to guide the eversion robot. At the distal end of the pipe, a 180\degree{} hollow U-bend was attached, with bending radii ranging from 40 mm down to 20 mm. These U-bends were 3D-printed from PLA and mounted interchangeably to create increasingly challenging confined trajectories.

Each eversion robot was inserted to a fixed starting point within the pipe and actuated at 3 kPa. Robots were allowed to evert until reaching the U-bend and were then kept pressurized. Some configurations successfully navigated through the entire U-bend and continued everting, while others became stuck. Trials in which the robot fully traversed the U-bend were labeled as successes, whereas those in which the robots were unable to complete the bend were labeled as failures.

For this experiment, three standard robots (no bands) and nine band-equipped robots (each with three constrictive bands), were tested. The bands were positioned 50 mm apart and located just before the bend. Of these nine robots, three used 10\% constrictive bands, three used 20\%, and three used 30\%. Each configuration was tested four times for each bending radius U-bend to ensure statistically significant results.

Between each trial, the robot’s initial orientation within the pipe was slightly varied so that the bands approached the U-bend at different angular and linear positions. This randomization simulates the uncertainty of real-world environments, in which the location and direction of bends are typically unknown, and in which the bands may encounter curvature at arbitrary points.

\subsubsection{Results}




The success rates for all tested configurations are summarized in Figure \ref{fig:Exp4}d. The standard robots with no bands were able to navigate 37.5 mm radius bends with a success rate of 100\%. When the bending radius was reduced to 35 mm, its success rate dropped to 58.3\%, and at 32.5 mm it further decreased to 33.3\%. Below this radius, the baseline robots consistently failed to complete the U-bend and were unable to continue everting.

In contrast, the robots equipped with stiffness-reducing bands were able to negotiate significantly tighter bends. The 10\% diameter-reduction design demonstrated the best performance among the configurations. These robots maintained excellent and highly consistent success rates down to a 30 mm bending radius. Even at 25 mm, the success rate remained high at 75\%. Consistent eversion only failed at a 20 mm bending radius, when the success rate dropped to 16.7\%.

However, increasing the diameter-reduction ratio from 10\% to 20\% and 30\% resulted in progressively lower success rates. We hypothesize that this trend arises from the increased internal pressure required to initiate eversion when larger reductions are implemented. Since all experiments were conducted at the same driving pressure (3 kPa) to enable fair comparison, the robots with higher-restriction bands likely everted less efficiently. As shown in Figure \ref{fig:Exp4}d, a 10\% reduction required only 0.62 kPa (SD=0.06) to initiate eversion, whereas the 20\% reduction required double the pressure, reaching 1.16 kPa (SD=0.19). This substantial increase in required pressure likely explains the reduced performance observed in higher-restriction designs.

\section{Case Study: Eversion in a Colon Phantom}

As a case study, we evaluated whether the constrictive bands enhance passive navigation in a colon-like environment. The goal of this study was to demonstrate the feasibility of the proposed passive steering strategy for colonoscopy, a medical procedure for which eversion robots are already being hailed as  innovative potential routes to combining diagnosis and therapy \cite{endotheranostics}. A standard eversion robot and a modified robot with integrated constrictive bands were deployed inside a silicone colon phantom while recording the motion of the eversion robot and corresponding pressure values.

\subsection{Materials and Methods}

The silicone colon phantom shown in Figure 5a was fabricated using Ecoflex 00-50 and dimensioned according to typical anatomical measurements as reported in the literature \cite{alazmani2016quantitative}. The sigmoid section of the colon phantom (the main part that is used in this study) was approximately 510 mm long with an internal diameter of 26 mm. Local surface undulations were introduced to replicate colonic haustral folds. The colon wall thickness was set to 3 mm. The selection of Ecoflex 00-50 was based on prior comparative studies indicating that its mechanical properties closely resemble those of human colon tissue, particularly in terms of stiffness and stretchability \cite{finocchiaro2023physical}.

To ensure high accuracy and geometric consistency, a modular fabrication approach was adopted. The colon geometry was split into anatomically defined segments corresponding to individual colonic regions (rectum, sigmoid, descending, transverse, ascending, and cecum), with segment dimensions derived from anatomical measurements described in the literature \cite{ourreview, alazmani2016quantitative}. Dedicated 3D-printed molds were designed for each segment, enabling the controlled reproduction of region-specific lengths, curvatures, and geometries. Physiologically relevant curvatures, such as the sigmoid loop, were explicitly embedded in the mold designs to reflect realistic anatomical configurations. Customized connection elements were developed to join adjacent segments, allowing assembly of a continuous colon phantom while preserving local anatomical features. This modular design enables reproducible fabrication while also allowing controlled variation of the global colon configuration when required.

\begin{figure}[ht]
  \centering
  \includegraphics[width=1\linewidth]{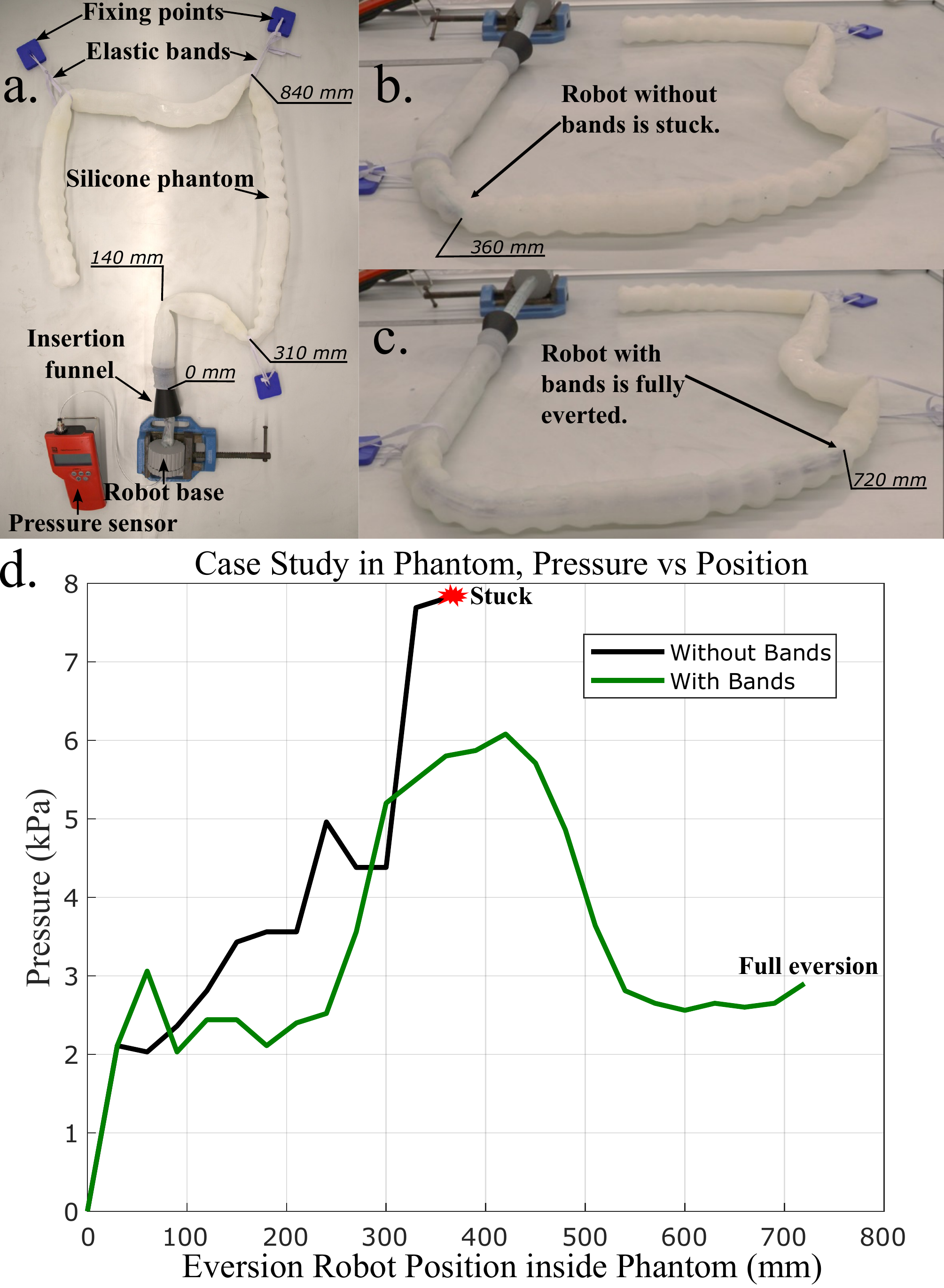}
  \caption{Case study using a colon phantom. a) A silicone phantom fixed to a whiteboard with elastic bands and magnets; the eversion robot is inserted using a funnel.  b) The standard eversion robot stalls at the 90\degree{} bend. c) The band-equipped robot successfully navigates the same bend and fully everts. d) Pressure–position plot showing that the standard robot gets stuck before the 400 mm mark, even at the maximum actuation pressure of 8 kPa, while the band-equipped robot completes the bend at a maximum actuation pressure of just over 6 kPa.}
  \label{CaseStudy}
\end{figure}

The phantom was positioned to create a sharp 90\degree{} bend in the sigmoid region (Figures~\ref{CaseStudy}b and c). Normally sigmoid colon can have a smaller bending radius \cite{alqarni2024experimental}. Although simplified, our configuration provided a controlled yet challenging navigation environment. Owing to the phantom’s variable shape, the bending radius varied from approximately 20 mm to 80 mm. The phantom was secured to a planar surface using elastic bands and magnets, simulating mesocolon elasticity, while the rectal end was fixed using an additional magnet. To avoid compressive blockage of the eversion robot at the rectum, an insertion funnel (with an outer diameter of 30 mm at the phantom anus) guided the robot into the phantom.

The eversion robot base was clamped to the board using a vise. Internal pressure was regulated via an SMC ITV2050 pressure regulator and the output of the regulator was fine-controlled with a handheld valve. Internal air pressure within the robot was measured using an RS Pro digital pressure indicator connected to the internal channel at the robot base. Whenever progress stalled, the pressure was increased gradually up to a maximum of 8 kPa, with the aim of restarting the robot's motion.  During the experiments, the inner pressure of the robot stayed always below 8 kPa to avoid seal rupture.

Two robot configurations were tested: a standard eversion robot and a robot equipped with three constrictive bands, each producing a 10\% diameter reduction and spaced 50 mm apart upstream of the bend. Both robots were 20 mm in diameter to match typical clinical colonoscope dimensions \cite{ourreview}, and approximately 800 mm in length, long enough to go beyond the 90\degree{} bend in the sigmoid region. Comparable bending angle navigation challenges have also been investigated in related studies \cite{borvorntanajanya2025pneumatic,shi2025design}. Full colon traversal was not evaluated in this study; the focus was on evaluating the navigation across the 90\degree{} bend. The eversion robots used in the experimental study were painted black to improve visibility through the semi-translucent phantom walls.

To actuate the robots, the pressure was gradually increased as the level of path tortuousness increased. Motion was recorded using two synchronized cameras (top and side views). Pressure readings were aligned with robot positions, logged at approximately every 30 mm of growth. Results are plotted in Figure \ref{CaseStudy}d.

\subsection{Results}

The standard robot (without bands) initially everted smoothly, traversing the straight segments with ease and requiring between 2 kPa and 4 kPa actuation pressure. However, the pressure demand increased sharply when approaching the 90-degree-bend at the 360 mm insertion depth position (reaching 8 kPa), at which point the robot got stuck and was unable to continue (Figure~\ref{CaseStudy}b).

The robot equipped with constrictive bands also passed the straight sections under pressures of between 2 kPa and 4 kPa. As it reached the bend, pressure increased to nearly 6 kPa, yet the robot successfully negotiated the 90-degree-bend and fully everted to its full length. This case study therefore confirms that the constrictive bands do indeed improve passive navigation under colon-like conditions.

Importantly, the colon phantom presented a more challenging environment than the rigid-tube tests (of Section IV, E). The variable diameter and elasticity of the phantom led to complex environmental dynamics. Unlike rigid pathways, the compliant phantom elongates before transmitting full reaction forces, creating local indentations that can trap the robot's tip. Navigating in this phantom without active navigation represents a significant challenge and demonstrates the key capability of the proposed design.

\section{Discussion and Conclusions}

This paper introduces a fundamental structural modification for eversion robots that significantly enhances their passive navigation capabilities. When an application requires path following in torturous environments the proposed method offers a simple, and highly effective solution: constrictive bands, fabricated from the same TPU material as the robot and integrated during manufacturing, locally reducing the robot diameter. Through mathematical modeling and experimental validation, we demonstrated that these constraints systematically reduce the bending stiffness at the robot tip and promote controlled buckling at predetermined locations. As a result, these robots become substantially more capable of conforming to sharper environmental turns than a standard eversion robot.

The robot body, with its constrictive bands, is modeled and analyzed using the proposed Cosserat rod formulation. The results indicate that the bands not only modify the global deformation shape but also lead to a substantial reduction in the effective elastic modulus within the diameter-reduced segments. This reduction originates from local slack and wrinkling of the pressurized tube wall, which diminish the hoop membrane tension responsible for pressure-induced stiffening. A reduction of approximately $45\%$ in effective elastic modulus is observed in response to  a $10\%$ diameter reduction, increasing to over $60\%$ for greater reductions. These reductions in effective modulus translate directly into proportional decreases in bending stiffness, ultimately enabling significantly improved passive navigation.

Our results indicate that distributing several bands along the robot's length yields the most reliable performance in unknown environments, where the location of bends cannot be predicted. The bending stiffness-reduction effect accumulates across bands placed upstream of the bending point, and we found that the band positioned furthest from the eventual contact point contributes the most. With respect to the degree of diameter restriction, the trend is clear: larger reductions lead to greater decreases in bending stiffness. However, this benefit comes with a trade-off. Increasing the restriction ratio substantially raises the internal pressure required for eversion. In practice, we observed that lower diameter restrictions (e.g., 10\%) provide the most effective balance between improved bending performance and manageable eversion pressures, whereas higher diameter restrictions (e.g., 50\%) become impractical due to excessive pressure demands.

Another important finding concerns band placement relative to the bend. A band located slightly upstream of the bending point (e.g., 20 mm before) is more effective than a band positioned exactly at the point of curvature. This introduces an element of randomness in unknown environments. However, this effect can be mitigated by increasing the frequency of bands along the robot body, thereby ensuring that at least one band is optimally positioned regardless of where bending occurs.

Although the rigid tube experiments involved sharper nominal bending radii, the colon phantom in the case study presented a substantially more challenging environment. The phantom’s compliant nature and variable diameter introduced sophisticated environmental dynamics that were absent in rigid test setups. Unlike rigid pathways, the elastic phantom deforms and elongates before fully transmitting reaction forces, leading to the formation of local indentations that can trap the robot. Successfully navigating a 90\degree{} bend under these conditions, under which the effective bending radius varied continuously between 20 mm and 80 mm and no active steering was employed, poses a significant challenge for passive navigation. The robot’s ability to traverse this environment therefore highlights the robustness and effectiveness of the proposed constrictive band design.

We also encountered practical challenges during fabrication. Robots with larger diameter-reduction ratios occasionally exhibited unintended pre-bending caused by slight asymmetries during ultrasonic welding. Re-welding from the opposite side reduced this issue, but the underlying problem highlights the sensitivity of the system to small manufacturing inconsistencies. As the robot performance increasingly depends on precise geometry, improving fabrication repeatability becomes essential. Future work will focus on machine-assisted manufacturing techniques to reduce human-error and ensure consistent prototype quality.

\section*{DISCLAIMER}

The first draft of Figure \ref{fig1}a was generated using ChatGPT, based on the prompt: “generate an image of an eversion robot with constrictive bands navigating an environment similar to the colon,” along with the abstract of the paper. Views and opinions expressed are however those of the author(s) only and do not necessarily reflect those of the European Union or the European Research Council Executive Agency. Neither the European Union nor the granting authority can be held responsible for them.

\section*{ACKNOWLEDGMENT}

The authors thank Mish Toszeghi for his editorial help, and our reviewers for their constructive comments.

\bibliographystyle{IEEEtran}
\bibliography{references.bib}

\newpage

 





\end{document}